# Randomized Memetic Artificial Bee Colony Algorithm

Sandeep Kumar[1], Dr. Vivek Kumar Sharma[2] and Rajani Kumari[3]

[1,2 & 3] Faculty of Engineering and Technology, Jagannath University,
Chaksu, Jaipur, Rajasthan 303901, India

**Abstract:** *Artificial Bee Colony (ABC) optimization algorithm is one of the recent population based probabilistic approach developed for global optimization. ABC is simple and has been showed significant improvement over other Nature Inspired Algorithms (NIAs) when tested over some standard benchmark functions and for some complex real world optimization problems. Memetic Algorithms also become one of the key methodologies to solve the very large and complex real-world optimization problems. The solution search equation of Memetic ABC is based on Golden Section Search and an arbitrary value which tries to balance exploration and exploitation of search space. But still there are some chances to skip the exact solution due to its step size. In order to balance between diversification and intensification capability of the Memetic ABC, it is randomized the step size in Memetic ABC. The proposed algorithm is named as Randomized Memetic ABC (RMABC). In RMABC, new solutions are generated nearby the best so far solution and it helps to increase the exploitation capability of Memetic ABC. The experiments on some test problems of different complexities and one well known engineering optimization application show that the proposed algorithm outperforms over Memetic ABC (MeABC) and some other variant of ABC algorithm(like Gbest guided ABC (GABC),Hooke Jeeves ABC (HJABC), Best-So-Far ABC (BSFABC) and Modified ABC (MABC) in case of almost all the problems.*

## 1. INTRODUCTION

Optimization is a field of research with large number of applications in almost each and every area of science, management and engineering, where mathematical modeling is used. Global optimization is a very challenging task due to diversity of objective functions and complex real world optimization problems as they may vary unimodel to multimodal, linear to highly non-linear and irregularities [1].

The growing trend of apprehensive informatics is to use swarm intelligence to search solutions to challenging optimization problems and real-world applications. A classical example is the evolution of particle swarm optimization (PSO) that may be recognized as a representative milestone in using swarm intelligence or collective intelligence [2]. One more, good example is the ant colony optimization algorithm which shows the collective intelligent behavior of social insects [3]. These strategies are meta-heuristic and swarm-based, still they share some common characteristics with genetic algorithms [2], [4], [5], but these methods are much simpler as they do not use mutation or crossover operators. Genetic algorithms are population-based, biology inspired algorithm, which was based on the evolution theory and survival of the fittest [6]. Similarly, PSO is also population-based, but it was inspired by the intelligent behavior of swarm like fish and birds, and therefore falls into the different class from genetic algorithms. Swarm-based or population-based algorithms fall in category of nature-inspired meta-heuristic algorithms [7]-[11] with a great range of applications areas [12]-[17]. Since the inception of swarm intelligence strategies such as PSO in the early 1990s, more than a dozen of new meta-heuristic algorithms have been developed [18]-[29], and these algorithms have been applied to almost every field of optimization, engineering, design, image, scheduling, data mining, machine intelligence etc.

The intelligent foraging behavior, learning capability, memorizing power and information sharing properties of honey bees have recently been one of the most interesting research areas in swarm intelligence. Research and studies on honey bees are setting a new trend in the literature during the last decade [30]. Artificial bee colony (ABC) algorithm was proposed by D. Karaboga in 2005 [31]. It is an optimization algorithm based on particular intelligent behavior of honey bee swarms. Since its inception it has been modified and compared with other meta-heuristics like genetic algorithm, particle swarm optimization (PSO), differential evolution (DE), and evolutionary algorithms[32], [33] on standard benchmark functions. It also has been used for designing of digital IIR filters [34], for training of feed forward neural network [35], to solve the leaf-constrained minimum spanning tree problem [36], for optimization of distribution network configuration[37], For optimization of truss structure [38] and for inverse analysis of structural parameter [39].

The rest of the paper is organized as follow: section 2 discuss original ABC algorithm in detail. Next section contains introduction to Memetic algorithms. Section 4 outlines proposed algorithm followed by experimental setup, results and discussion. Section 6 contains conclusions followed by references.

## 2. ARTIFICIAL BEE COLONY ALGORITHM

Artificial Bee Colony (ABC) Algorithm is inspired by the intuitive food foraging behavior of honey bee insects. Honey bee swarm is one of the most insightful insect exists in nature; it shows collective intelligent behavior while searching the food. The honey bee swarm has a number of good features like bees can interchange the information, can remember the environmental conditions, can store and share the information and take decisions based these observations. According to changes in the environment, the honey bee swarm updates itself, allocate the tasks actively and proceed further by social learning and teaching. This intelligent behavior of honey bees inspires researchers and scientists to simulate the intelligent food foraging behavior of the honey bee swarm.

### 2.1 Analogy of Artificial Bee Colony Algorithm with Bee

The algorithm suggested by D. Karaboga [31] is collection of three major elements: employed bees, unemployed bees and food sources. The employed bees are associated with a good quality food source. Employed bees have good knowledge about existing food source. Exploitation of food sources completed by employed bees. When a food source rejected employed bee turn to be as unemployed. The idle foragers are bees acquiring no information about food sources and searching for a new food source to accomplish it. Unemployed bees can be classified in two categories: scout bees and onlooker bees. Scout bees randomly search for new food sources neighboring the hive. Onlooker bees monitor the waggle dance in hive, in order to select a food source for the purpose of exploitation. The third element is the good food sources nearby the beehive.

Comparatively in the optimization context, the number of food sources (that is the employed or onlooker bees) in ABC algorithm, is commensurate to the number of solutions in the population. Moreover, the location of a food source indicates the position of an encouraging solution to the optimization problem, after all the virtue of nectar of a food source represents the fitness cost (quality) of the correlated solution.

### 2.2 Phases of Artificial Bee Colony Algorithm

The search process of ABC has three major steps [31]:

- Send the employed bees to a food source and estimate their nectar quality;
- Onlooker bees select the food sources based on information collected from employed bees and estimate their nectar quality;
- Determine the scout bees and employ them on possible food sources for exploitation.

The position of the food sources are arbitrarily selected by the bees at the initial stage and their nectar qualities are measured. The employed bees then communicate the nectar information of the sources with the onlooker bees waiting at the dance area within the hive. After exchanging this information, each employed bee returns to the food source checked during the previous cycle, as the position of the food source had been recalled and then selects new food source using its perceived information in the neighborhood of the present food source. In the last phase, an onlooker bee uses the information retrieved from the employed bees at the dance area to select a good food source. The chances for the food sources to be elected boosts with boost in its quality of nectar. Hence, the employed bee with information of a food source with the highest quality of nectar employs the onlookers to that food source. It eventually chooses another food source close by the one presently in her memory depending on perceived information. A new food source is arbitrarily generated by a scout bee to replace the one abandoned by the onlooker bees. This complete search process of ABC algorithm could be outlined in Fig. 1 as follows [31]:

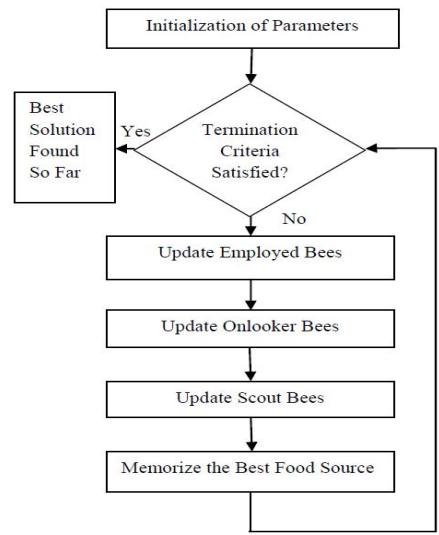

**Figure 1:** Phases of Artificial Bee Colony Algorithm

#### 2.2.1 Initialization of Swarm

The ABC algorithm has three main parameters: the number of food sources (population), the number of attempt after which a food source is treated to be jilted (limit) and the criteria for termination (maximum number of cycle). In the initial ABC proposed by D. Karaboga [31], initial population (SN) divided in two equal parts (employed bees (SN/2) or onlooker bees (SN/2)); the number of food sources was equal to the employed bees (SN/2). Initially it consider an evenly dealt swarm of food sources (SN), where each food source $x_i$ (i = 1, 2 ...SN) is a vector of D-dimension. Each food source is evaluated using following method [40]:

$$x_{ij} = x_{\min j} + rand[0,1](x_{\max j} - x_{\min j}) \quad (1)$$

Where

– rand[0,1] is a function that arbitrarily generates an evenly distributed random number in range [0,1].

### 2.2.2 Employed Bee
Employed bees phase modify the existing solution according to information of individual experiences and the fitness value of the newly raised solution. New food source with higher fitness value take over the existing one. The position of $i^{th}$ candidate in $j^{th}$ dimension updated during this phase as shown below [40]:

$$v_{ij} = x_{ij} + \phi(x_{ij} - x_{kj}) \quad (2)$$

Where
– $\phi(x_{ij} - x_{kj})$ is known as step size, k ∈ {1, 2, ..., SN}, j ∈ {1, 2, ...,D} are two randomly chosen indices. k ≠ i ensure that step size has some indicative improvement.

### 2.2.3 Onlooker Bee
The number of food sources for onlooker bee is same as the employed. All along this phase all employed bee exchange fitness information about new food sources with onlooker bees. Onlooker bees estimate the probability of selection for each food source generated by the employed bee. The onlooker bee select best fittest food source. There may be some other methods for calculation of probability, but it is required to include fitness. In this paper probability of each food source is decided using its fitness as follow [40]:

$$P_{ij} = \frac{fit_i}{\sum_{i=1}^{SN} fit_i} \quad (3)$$

### 2.2.4 Scout Bee Phase
If the location of a food source is not modified for a predefined number of evaluations, then the food source is assumed to be neglected and scout bees phase is initialized. All along this phase the bee associated with the neglected food source converted into scout bee and the food source is replaced by the arbitrarily chosen food source inside the search space. In ABC, the pretended number of cycles plays an important role in form of control parameter which is called limit for rejection of food sources. Now the scout bees replace the abandoned food source with new one using following equation [40].

$$x_{ij} = x_{\min j} + rand[0,1](x_{\max j} - x_{\min j}) \quad (4)$$

Based on the above description, it is clear that in ABC search process there are three main control parameters: the number of food sources SN, the limit and the maximum number of cycles. The steps of ABC algorithm are outlined as follow [40]:

---

Algorithm 1: Artificial Bee Colony Algorithm
Initialize all parameters;
Repeat while Termination criteria is not fulfilled
  Step 1: Employed bee phase for calculating new food sources.
  Step 2: Onlooker bees phase for updating location the food sources based on their amount of nectar.
  Step 3: Scout bee phase in order to search new food sources in place of rejected food sources.
  Step 4: Remember the best food source identified so far.
End of while
Output the best solution found so far.

---

## 3. MEMETIC ALGORITHMS
Memetic Algorithms (MAs) was the name introduced by P.A. Moscato [41] to a class of stochastic global search techniques that, commonly speaking, combine within the framework of Evolutionary Algorithms (EAs) the benefits of problem-specific local search heuristics and multi-agent systems. In ethnic advancement processes, information is processed and magnified by the communicating parts, it is not only transmitted unaltered between entities. This enhancement is realized in MAs by assimilating heuristics, approximation algorithms, metaheuristics, local search techniques, particular recombination operators, truncated exact search methods, etc. In aspect, almost all MAs can be illustrate as a search technique in which a population of optimizing operator co-operate and contest. MAs have been successfully enforced to a wide range of domains that envelop problems in combinatorial optimization, like E. Burke, J. Newall, R. Weare [42] applied memetic algorithm for university exam timetabling. R. Carr, W. Hart, N. Krasnogor, E. Burke, J. Hirst, J. Smith [43] used a memetic evolutionary algorithm for alignment of protein structures. R. Cheng, M. Gen [44] scheduled parallel machine using memetic algorithms. C. Fleurent, J. Ferland [45] developed a graph coloring algorithm with hybridization of Genetic algorithm. It is also applied for travelling salesman problem [46], back routing in telecommunication [47], bin packing [48], VLSI floor planning [49] continuous optimization [50],[51], dynamic optimization [52], multi-objective optimization [53].

Memetic Algorithm play an important role in evolutionary computing as it provide better exploitation capability of search space in local search. Application area of memetic algorithms continuously expending after its inception. Y. Wang et al [54] developed a memetic algorithm for the maximum diversity problem based on tabu search. X. Xue et al [55] Optimize ontology alignment with help of Memetic Algorithm based on partial reference alignment. O. Chertov and Dan Tavrov [56] proposed a memetic algorithm for solution of the task of providing group anonymity. J. C. Bansal et al [57] incorporated Memetic search in artificial bee colony algorithm. F. Kang et al [58] developed a new memetic algorithm HJABC. I. Fister et al [59] proposed a memetic artificial bee colony algorithm for large-scale global optimization.

Memetic algorithm proposed by Kang et al. [58] incorporates Hooke Jeeves [60] local search method in Artificial Bee Colony algorithm. HJABC is an algorithm with intensification search based on the Hooke Jeeves pattern search and the ABC. In the HJABC, two modification are proposed, one is the fitness ($Fit_i$) calculation function of basic ABC is changed and calculated by equation (5) and another is that a Hooke-Jeeves local search is incorporated with the basic ABC.

$$fit_i = 2 - SP + \frac{2(SP-1)(p_i-1)}{(NP-1)} \quad (5)$$

Here $p_i$ is the position of the solution in the whole population after ranking, SP ∈ [1.0, 2.0] is the selection pressure. A medium value of SP = 1.5 can be a good choice and NP is the number of solutions.

HJABC contains iterations of exploratory move and pattern move to search optimum result of problem. Exploratory move consider one variable at a time in order to decide appropriate direction of search. The next step is pattern search to speed up search in decided direction by exploratory move. These two steps repeated until the termination criteria meet. The Hooke-Jeeves pattern move is a contentious attempt of the algorithms for the exploitation of promising search directions as it collect information from previous successful search iteration. Algorithm 2 outlines major steps of HJABC [58].

---

**Algorithm 2: Hooke-Jeeves ABC Algorithm**
Initialize the all the parameters
  Population of solutions $x_i$, $l =1,…,NS$.
  Evaluate the population for, $cycle$=1, $c$=0.
  Remember the best solution $x_{best}$ and set $x_{best1} = x_{best}$
**Repeat**
*Phase I: Exploration*
  Generate new solutions $v_i$ for the employed bees by using (2) and evaluate them.
  Apply the greedy selection approach for the employed bees.
  Grade the population and quantify the fitness by (5)
  Quantify the probability $p_i$ for the solutions $x_i$ by (3).
  Generate the new solutions $v_i$ for the onlooker bees from the solutions selected based on $p_i$ and assess them.
  Apply the greedy selection approach for the onlooker bees.
  Determine the abandoned solution for the scout, if exists, and replace it with a new randomly produced solution $x_i$.
  Retain the best solution $x_{best}$ attained so far.
*Phase II: Exploitation*
If (($cycle$ mod $interval$) = 0),
  Quantify step size $\delta_j$ of $j^{th}$ dimension m number of selected solutions in Hooke-Jeeves according to
    $\delta_j = 0.1 * \frac{\sum_{i=1}^{m}(X'_{ij} - x_{bestj})}{m}$.
  Call modified Hooke-jeeves with $x_{best}$ as the base point until $\delta_d < \varepsilon$ and the obtained point is $x_{best2}$.
  If ($f(x_{best2}) \leq f(x_{best})$)
    Substitute the solution in the central position after ranking by $x_{best1}$ and set $x_{best1} = x_{best2}$
  If ($f(x_{best1}) < f(x_{best})$)
    set $x_{best} = x_{best1}$ and $k$=0, else set $k$=$k$+1;
  Set $cycle$=$cycle$+1.
*Phase III: Intensification search*
  If ($c > counter$)
    Perform intensification search by using modified Hooke- Jeeves.
**Until termination criteria is met.**

---

The MeABC algorithm proposed by J. C. Bansal et al [57] inspired by Golden Section Search (GSS) [61]. In MeABC only the best particle of the current swarm updates itself in its proximity. Original GSS method does not use any gradient information of the function to finds the optima of a uni-modal continuous function. GSS processes the interval [a = −1.2, b = 1.2] and initiates two intermediate points:

$$F_1 = b − (b − a) \times \psi, \quad (6)$$
$$F_2 = a + (b − a) \times \psi, \quad (7)$$

Here $\psi$ = 0.618 is the golden ratio.
The MeABC contains four steps:
 – employed bee phase,
 – onlooker bee phase,
 – scout bee phase and
 – memetic search phase

First three steps are similar to the original ABC [31] except updating of position of an individual. It updates position as per the following equation.

$$x'_{ij} = x_{ij} + \phi_{ij}(x_{ij} − x_{kj}) + \psi_{ij}(x_{bestj} − x_{ij})$$
$$(8)$$

Where $\Psi_{ij}$ is a random number in interval [0, C], here C is a positive constant. The detailed working of the MeABC [57] explained in Algorithm 3.

---

**Algorithm 3: MeABC Algorithm**
Initialize the parameters.
Repeat while termination criteria meet
  Step I: Produce new solutions for the employed bees
  Step II: Produce the new solutions for the onlookers from the solutions selected depending on pi and evaluate them.
  Step III: Determine the abandoned solution for the scout, if exists, and replace it with a new randomly produced solution.
  Step IV: Memorize the best solution achieved so far.
  Phase V: Apply Memetic search as given by [57]

---

## 4. RANDOMIZED MEMETIC ARTIFICIAL BEE COLONY ALGORITHM

J. C. Bansal, H. Sharma and S. S. Jadon [17] indicate some drawbacks of population based stochastic algorithms. It is observed that the early convergence or stagnation occurs in ABC. The position update in ABC takes place with help of equation

$$v_{ij} = x_{ij} + \phi(x_{ij} − x_{kj}) \quad (9)$$

After a certain number of iterations, normally all probable solutions work within a very small neighborhood. In this case the variation $x_{ij} − x_{kj}$ becomes very small and so the enhancement in the position becomes inconsequential. This event is known as the stagnation or premature convergence if the global optimal solution is not present in this small neighborhood. Any population based algorithm is observed as an adequate algorithm if it is swift in convergence and able to explore the large area of the search space. Especially, if a population based stochastic algorithm is able to balance between exploration and exploitation of the search space, then the algorithm is regarded an adequate algorithm. From this point of view, original ABC is not an efficient algorithm [40].

D. Karaboga and B. Akay [40] analyzed the distinct variants of ABC for global optimization and conclude that ABC shows a poor performance and remains incompetent in exploring the search space. The problem of early convergence and stagnation is a matter of severe consideration for evolution a relatively more efficient ABC algorithm. The proposed randomized memetic ABC introduce two new control parameters in memetic search phase.

The proposed RMABC contains four phases. First three phases: employed bee phase, onlooker bee phase and scout bee phase are similar to original MeABC [57]. During fourth phase (memetic search phase), it introduce two new parameters in order to strengthen local search process. Golden section search [61] process controls the step size during updating of best individual in current swarm. In RMABC the GSS processes the interval [a = -1.2, b = 1.2] and generate two intermediate points:

$f_1 = \phi_1 * (b - (b-a) * \Psi)$ (10)
$f_2 = \phi_2 * (a + (b-a) * \Psi)$ (11)
Where
- $\phi_1$ is a positive constant in interval [0, 1],
- $\phi_2$ is a Negative constant in interval [-1, 0] and
- $\psi = 0.618$ is the golden ratio.

Opposite sign of $\phi_1$ and $\phi_2$ make it convenient to explore local search space. Randomly generated value of $\phi_1$ and $\phi_2$ in specified range avoid stagnation and premature convergence. The proposed population based stochastic algorithm tries to balance between intensification and diversification of the large search space. It arbitrarily navigates almost complete search space during local search phase and inching ahead towards global optimum. In RMABC only the best solution of the current population modify itself in its proximity. Here position update process also modified as shown in equation (8). This modified memetic search phase applied after scout bee phase of original ABC not during local search phase. Algorithm 4 outlines RMABC in detail with description of each step. The symbol and parameters all are taken as it was in original memetic artificial bee colony algorithm [57].

**Algorithm 4: Randomized Memetic ABC Algorithm**
Initialize the population of solutions $x_i$, $i = 1, 2, ..., NS$ and Evaluate the population using.
$$x_{ij} = x_{\min j} + rand[0,1](x_{\max j} - x_{\min j})$$
**Repeat** while termination criteria meet
  Generate and evaluate new solutions $v_{ij}$ ($i^{th}$ solution in $j^{th}$ direction) for the employed bees by using
$$v_{ij} = x_{ij} + \phi(x_{ij} - x_{kj})$$
  Apply the greedy selection approach for the employed bees.
  Grade the population and quantify the fitness by
  if (fun_val >= 0)
    Then $fit_i = \frac{1}{(2*fun\_val+1)}$
  else
    $fit_i = (1 + fabs(\frac{1}{fun\_val}))$
  Quantify the probability $p_i$ for the solutions $x_i$ with fitness $fit_i$ by
$$P_{ij} = \frac{fit_i}{\sum_{i=1}^{SN} fit_i}$$
  Generate the new solutions $x_i$ for the onlooker bees from the solutions selected based on $p_i$ and assess them using
$$x'_{ij} = x_{ij} + \phi_{ij}(x_{ij} - x_{kj}) + \psi_{ij}(x_{bestj} - x_{ij})$$
  Determine the discarded solution for the scout, if exists, and replace it with a new randomly produced solution $x_i$.
  Retain the best solution $X_{best}$ attained so far.
  Memetic search phase

  Initialize a= -1.2 and b = 1.2
  Repeat while (|a-b|<ϵ)
    Compute $f_1 = \phi_1*(b - (b-a)*\Psi)$,  Where $\phi_1 \in [0, 1]$
              $f_2 = \phi_2*(a + (b-a)*\Psi)$, Where $\phi_2 \in [-1,0]$
    Generate two new solutions $X_{new1}$ and $X_{new2}$ using $f_1$ and $f_2$ respectively according to MeABC
    Calculate $f(X_{new1})$ and $f(X_{new2})$ for objective function
    if ($f(X_{new1}) < f(X_{new2})$) then $b = f_2$
      if ($f(X_{new1}) < f(X_{best})$)
        then $X_{best} < X_{new1}$
    else
      $a = f_1$
      if ($f(X_{new2}) < f(X_{best})$)
        then $X_{best} < X_{new2}$

**Table 1**: Test Problems

| Test Problem | Objective Function | Search Range | Optimum value | D | Acceptable Error |
|---|---|---|---|---|---|
| Zakharov | $f_1(x) = \sum_{i=1}^{D} x_i^2 + (\sum_{i=1}^{D} \frac{ix_i}{2})^2 + (\sum_{i=1}^{D} \frac{ix_i}{2})^4$ | [-5.12, 5.12] | $f(0) = 0$ | 30 | 1.0E-02 |
| Salomon Problem | $f_2(x) = 1 - \cos(2\pi\sqrt{\sum_{i=1}^{D} x_i^2}) + 0.1(\sqrt{\sum_{i=1}^{D} x_i^2})$ | [-100, 100] | $f(0) = 0$ | 30 | 1.0E-01 |
| Sum of different powers | $f_3(x) = \sum_{i=1}^{D} |x_i|^{i+1}$ | [-1, 1] | $f(0) = 0$ | 30 | 1.0E-05 |
| Levy montalvo 1 | $f_4(x) = \frac{\pi}{D}(10\sin^2(\pi y_1) + \sum_{i=1}^{D-1}(y_i-1)^2(1+10\sin^2(\pi y_{i+1})) + (y_D-1)^2)$, Where $y_i = 1 + \frac{1}{4}(x_i+1)$ | [-10, 10] | $f(-1) = 0$ | 30 | 1.0E-05 |
| Levy montalvo 2 | $f_5(x) = 0.1(\sin^2(3\pi x_1) + \sum_{i=1}^{D-1}(x_i-1)^2(1+\sin^2(3\pi x_{i+1})) + (x_D-1)^2(1+\sin^2(2\pi x_D)))$ | [-5, 5] | $f(1) = 0$ | 30 | 1.0E-05 |
| Beale function | $f_6(x) = (1.5 - x_1(1-x_2))^2 + (2.25 - x_1(1-x_2^2))^2 + (2.625 - x_1(1-x_2^3))^2$ | [-4.5, 4.5] | $f(3, 0.5) = 0$ | 2 | 1.0E-05 |
| Colville function | $f_7(x) = 100(x_2 - x_1^2)^2 + (1-x_1)^2 + 90(x_4 - x_3^2)^2 + (1-x_3)^2 + 10.1[(x_2-1)^2 + (x_4-1)^2] + 19.8(x_2-1)(x_4-1)$ | [-10, 10] | $f(1) = 0$ | 4 | 1.0E-05 |
| Kowalik function | $f_8(x) = \sum_{i=1}^{11}(a_i - \frac{x_1(b_i^2+b_ix_2)}{b_i^2+b_ix_3+x_4})^2$ | [-5, 5] | $f(0.1928, 0.1908, 0.1231, 0.1357) = 3.07E-04$ | 4 | 1.0E-05 |
| Shifted Rosenbrock | $f_9(x) = \sum_{i=1}^{D-1}(100(z_i^2-z_{i+1})^2 + (z_i-1)^2 + f_{bias})$, $z = x - o + 1, x = [x_1,x_2,...x_D], o = [o_1,o_2,......o_D]$ | [-100, 100] | $f(o) = f_{bias} = 390$ | 10 | 1.0E-01 |

## 5. Experimental results and discussion
### 5.1 Test problems under consideration
In order to analyze the performance of RMABC, 9 different global optimization problems ($f_1$ to $f_9$) are selected (listed in Table 1). These are continuous optimization problems and have different degrees of complexity and multimodality. Test problems $f_1$ –$f_8$ are taken from [62] and test problems $f_9$ is taken from [63] with the associated offset values.

### 5.2 Experimental setting
To prove the efficiency of RMABC, it is compared with ABC and recent variants of ABC named MeABC [57], Gbestguided ABC (GABC) [64], Best-So-Far ABC (BSFABC) [65], HJABC [58] and Modified ABC (MABC) [66]. To test Randomized memetic ABC (RMABC) over considered problems, following experimental setting is adopted:
  − Colony size NP = 50 [57],
  − $\phi_{ij}$ = rand[−1, 1],
  − Number of food sources SN = NP/2,
  − limit = 1500 [57],
  − The stopping touchstone is either maximum number of function evaluations (which is set to be 200000) is reached or the acceptable error (outlined in Table 1) has been achieved,
  − The number of simulations/run =100,
  − Value of termination criteria in memetic search phase is set to be $\varrho = 0.01$.

Parameter settings for the algorithms ABC, MeABC, GABC, BSFABC, HJABC and MABC are similar to their original research papers.

### 5.3 Results Comparison
Numerical results of RMABC with experimental setting as per subsection 5.2 are outlined in Table 2. Table 2 show the comparison of results based on standard deviation (SD), mean error (ME), average function evaluations (AFE) and success rate (SR) are reported. Table 2 shows that most of the time RMABC outperforms in terms of efficiency (with less number of function evaluations) and accuracy as compare to other considered algorithms.

**Table 2:** Comparison of the results of test problems

| Test Function | Measure | RMABC | MeABC | ABC | HJABC | GABC | BSFABC | MABC |
|---|---|---|---|---|---|---|---|---|
| $f_1$ | SD | 3.55E-04 | 5.10E-04 | 1.52E+01 | 5.49E-02 | 1.89E+01 | 1.22E+01 | 1.02E-01 |
|  | ME | 9.68E-03 | 9.58E-03 | 9.73E+01 | 8.94E-02 | 9.73E+01 | 8.49E+01 | 1.46E-01 |
|  | AFE | 48657.79 | 94564.52 | 200000 | 198146 | 200000.01 | 200000 | 200005.52 |
|  | SR | 100 | 100 | 0 | 4 | 0 | 0 | 0 |
| $f_2$ | SD | 2.28E-02 | 3.28E-02 | 6.25E-02 | 3.45E-02 | 3.38E-02 | 6.58E-02 | 3.46E-02 |
|  | ME | 9.04E-01 | 9.24E-01 | 9.56E-01 | 9.12E-01 | 9.32E-01 | 9.53E-01 | 9.31E-01 |
|  | AFE | 23467.73 | 18209.52 | 149071.35 | 20266.59 | 75922.34 | 184747.81 | 27739.5 |
|  | SR | 100 | 100 | 68 | 100 | 98 | 74 | 100 |
| $f_3$ | SD | 2.75E-06 | 2.99E-06 | 2.79E-06 | 2.94E-06 | 2.73E-06 | 2.54E-06 | 1.99E-06 |
|  | ME | 6.24E-06 | 5.24E-06 | 5.84E-06 | 5.44E-06 | 5.60E-06 | 5.92E-06 | 7.51E-06 |
|  | AFE | 1864.32 | 4738.72 | 15619.5 | 4989.58 | 9290.5 | 14278 | 9422 |
|  | SR | 100 | 100 | 100 | 100 | 100 | 100 | 100 |
| $f_4$ | SD | 1.21E-06 | 7.97E-07 | 2.24E-06 | 6.05E-07 | 1.99E-06 | 2.41E-06 | 7.37E-07 |
|  | ME | 9.05E-06 | 9.16E-06 | 7.21E-06 | 9.24E-06 | 7.84E-06 | 6.98E-06 | 9.17E-06 |
|  | AFE | 10560.96 | 11770.12 | 19614.5 | 19214.65 | 13030.5 | 26863 | 22548.5 |
|  | SR | 100 | 100 | 100 | 100 | 100 | 100 | 100 |
| $f_5$ | SD | 9.41E-07 | 7.56E-07 | 2.13E-06 | 6.49E-07 | 1.83E-06 | 2.41E-06 | 8.14E-07 |
|  | ME | 9.06E-06 | 9.10E-06 | 7.35E-06 | 9.18E-06 | 8.10E-06 | 7.13E-06 | 9.06E-06 |
|  | AFE | 11261.3 | 13031.58 | 22016 | 17368.82 | 14283 | 28573.5 | 20985.5 |
|  | SR | 100 | 100 | 100 | 100 | 100 | 100 | 100 |
| $f_6$ | SD | 3.10E-06 | 2.90E-06 | 1.73E-06 | 2.99E-06 | 2.92.E-06 | 6.07E-05 | 3.06E-06 |
|  | ME | 4.94E-06 | 5.14E-06 | 8.58E-06 | 4.76E-06 | 5.14E-06 | 2.19E-05 | 5.24E-06 |
|  | AFE | 1594.35 | 2688.15 | 15768.28 | 4839.56 | 9344.1 | 50222.41 | 10082.84 |
|  | SR | 100 | 100 | 100 | 100 | 100 | 92 | 100 |
| $f_7$ | SD | 2.62E-03 | 2.42E-03 | 1.07E-01 | 2.52E-03 | 1.24E-02 | 2.99E-02 | 7.72E-03 |
|  | ME | 6.99E-03 | 7.03E-03 | 1.67E-01 | 7.14E-03 | 1.58E-02 | 2.18E-02 | 1.26E-02 |
|  | AFE | 29441.96 | 30813.41 | 198058.11 | 43566.59 | 154523.83 | 155548.21 | 144033.7 |
|  | SR | 100 | 100 | 2 | 100 | 42 | 47 | 54 |
| $f_8$ | SD | 1.98E-05 | 2.03E-05 | 7.32E-05 | 5.58E-05 | 3.57E-05 | 8.16E-05 | 6.84E-05 |
|  | ME | 8.30E-05 | 8.17E-05 | 1.69E-04 | 1.18E-04 | 9.27E-05 | 1.45E-04 | 1.90E-04 |
|  | AFE | 43472.89 | 47100.43 | 178355..83 | 127096.42 | 98389.57 | 140918.92 | 191449.61 |
|  | SR | 100 | 100 | 23 | 58 | 90 | 51 | 10 |
| $f_9$ | SD | 3.86E-02 | 7.83E-02 | 9.44E-01 | 7.24E-01 | 7.56E-02 | 5.63E+00 | 9.67E-01 |
|  | ME | 9.82E-02 | 1.03E-01 | 6.79E-01 | 5.79E-01 | 9.30E-02 | 2.96E+00 | 6.80E-01 |
|  | AFE | 98163.41 | 103949.03 | 175270.8 | 151927.05 | 100594.41 | 185221.92 | 163969.65 |
|  | SR | 91 | 97 | 24 | 46 | 93 | 13 | 39 |

Some more intensive analyses based on performance indices have been carried out for results of ABC and its variants in figure 2 to 8. Table 4 proves that the proposed method is superior to others considered algorithms. The proposed RMABC algorithm drastically reduces number of function evaluation for al problems. Entries shown in bold font in table 2 represents improved solutions. The proposed algorithm sometimes reduces number of AFE upto 40%. Here except function $f_2$, each time it improves in terms of AFE.

## 6. Applications of RMABC to Engineering Optimization Problem

To see the robustness of the proposed strategy, a well known engineering optimization problems, namely, Compression Spring [67] [68] is solved. The considered engineering optimization problem is described as follows:

### 6.1 Compression Spring

The considered engineering optimization application is compression spring problem [48], [56]. This problem minimizes the weight of a compression spring, subject to constraints of minimum bending, prune stress, deluge frequency, and limits on outside diameter and on design variables. There are three important design variables: the diameter of wire $x_1$, the diameter of the mean coil $x_2$, and the count of active coils $x_3$. Here it is presented in simplified manner. The mathematical formulation of this problem is outlined below:

$$x_1 \in \{1, 2, 3, \ldots, 70\} granularity_1$$

$$x_2 \in [0.6; 3], x_3 \in [0.207; 0.5] granularity_{0.001}$$

And four constraints

$$g_1 := \frac{8c_f F_{max} x_2}{\pi x_3^3} - S \leq 0, \quad g_2 := l_f - l_{max} \leq 0$$

$$g_3 := \sigma_p - \sigma_{pm} \leq 0, \quad g_4 := \sigma_w - \frac{F_{max} - F_p}{K} \leq 0$$

Where:

$$c_f = 1 + 0.75 \frac{x_3}{x_2 - x_3} + 0.615 \frac{x_3}{x_2}$$

$$F_{max} = 1000, S = 189000, l_f = \frac{F_{max}}{K} + 1.05(x_1 + 2)x_3$$

$$l_{max} = 14, \sigma_p = \frac{F_p}{K}, \sigma_{pm} = 6, F_p = 300$$

$$K = 11.5 \times 10^6 \frac{x_3^4}{8x_1 x_2^3}, \sigma_w = 1.25$$

And the function to be minimized is

$$f_{10}(X) = \pi^2 \frac{x_2 x_3^2 (x_1 + 2)}{4}$$

The best known solution is (7, 1.386599591, 0.292), which gives the fitness value $f = 2.6254$. Acceptable error for compression spring problem is $1.0E-04$.

**Table 3:** Comparison of the results of Compression spring problem ($f_{10}$)

| Algorithm | SD | ME | AFE | SR |
|---|---|---|---|---|
| RMABC | 4.21E-04 | 4.88E-04 | 38052.31 | 99 |
| MeABC | 2.37E-03 | 1.71E-03 | 123440.3 | 62 |
| ABC | 1.17E-02 | 1.36E-02 | 187602.32 | 10 |
| HJABC | 1.53E-03 | 1.17E-03 | 109737.22 | 70 |
| GABC | 9.50E-03 | 8.64E-03 | 189543.56 | 11 |
| BSFABC | 3.08E-03 | 3.02E-02 | 200031.13 | 0 |
| MABC | 6.59E-03 | 5.28E-03 | 181705.01 | 15 |

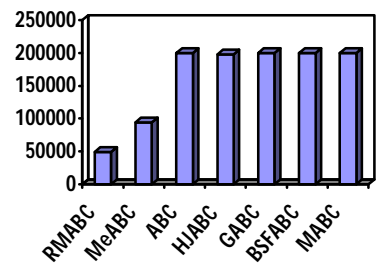

**Figure 2:** Comparison of AFE for function $f_1$

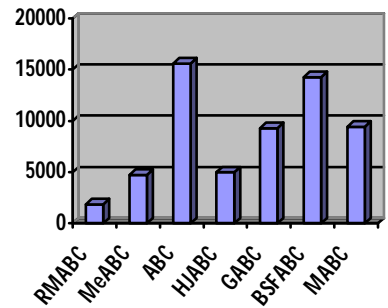

**Figure 3:** Comparison of AFE for function $f_3$

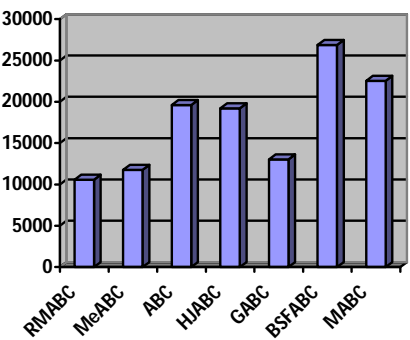

**Figure 4:** Comparison of AFE for function $f_4$

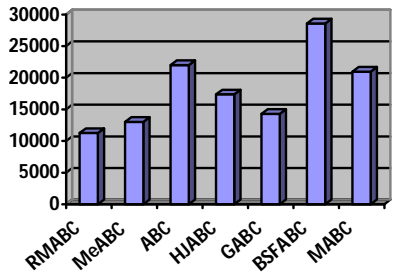

**Figure 5:** Comparison of AFE for function $f_5$

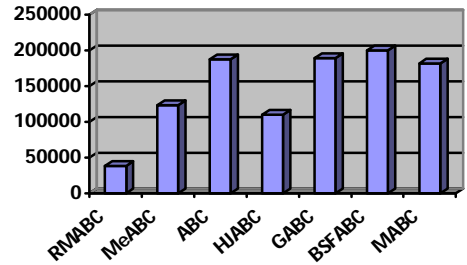

**Figure 8:** Comparison of AFE for function $f_{10}$

## 7. CONCLUSION

Here in this paper, two new parameters, $\phi_1$ and $\phi_2$ are introduced in memetic search phase of MeABC. Value of these parameters is arbitrarily chosen every time so obtained modified MeABC is named as Randomized Memetic ABC (RMABC). During its memetic search phase, Golden Section Search method is used for generating the new solutions nearby the best solution. Due to newly introduced parameter ($\phi_1$ and $\phi_2$) new solutions are evolved in the proximity the best solution. Further, the modified strategy is also applied to the recent variants of ABC, namely, GABC, HJABC, BSFABC and MABC. With the help of experiments over test problems and a well known engineering optimization application, it is shown that the inclusion of the proposed strategy in the memetic ABC improves the reliability, efficiency and accuracy as compare to their original version. Table 4 show that the proposed RMABC is able solves almost all the considered problems.

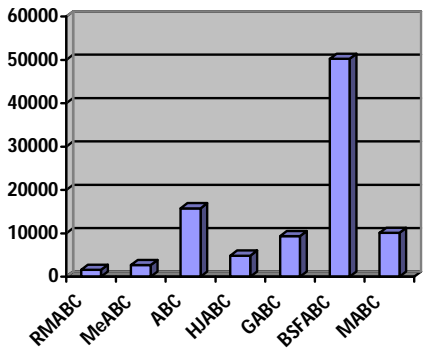

**Figure 6:** Comparison of AFE for function $f_6$

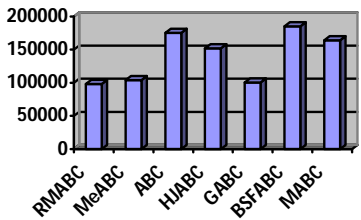

**Figure 7:** Comparison of AFE for function $f_9$

**Table 4:** Summary of table 2 and 3 outcome

| Function | RMABC Vs MeABC | RMABC Vs ABC | RMABC Vs HJABC | RMABC Vs GABC | RMABC Vs BSFABC | RMABC Vs MABC |
|---|---|---|---|---|---|---|
| $f_1$ | + | + | + | + | + | + |
| $f_2$ | + | + | + | + | + | + |
| $f_3$ | + | + | + | + | + | + |
| $f_4$ | + | + | + | + | + | + |
| $f_5$ | + | + | + | + | + | + |
| $f_6$ | + | + | + | + | + | + |
| $f_7$ | + | + | + | + | + | + |
| $f_8$ | + | + | + | + | + | + |
| $f_9$ | - | + | + | + | + | + |
| $f_{10}$ | + | + | + | + | + | + |
| Total number of + sign | 9 | 10 | 10 | 10 | 10 | 10 |

## References

[1] Georgieva and I. Jordanov, "Global optimization based on novel heuristics, low-discrepancy sequences and genetic algorithms," Eur. J. Oper. Res., vol. 196, pp. 413–422, 2009.

[2] J. Kennedy and R.C. Eberhart, "Particle swarm optimization," in: Proc. of IEEE International Conference on Neural Networks, Piscataway, NJ. pp. 1942-1948, 1995.

[3] M. Dorigo, "Optimization, Learning and Natural Algorithms," PhD thesis, Politecnico di Milano, Italy, 1992.

[4] W.J. Bell, "Searching Behaviour: The Behavioural Ecology of Finding Resources," Chapman & Hall, London, 1991.

[5] A.P. Engelbrecht, "Fundamentals of Computational Swarm Intelligence," Wiley, 2005.

[6] J. Holland, "Adaptation in Natural and Artificial systems," University of Michigan Press, Ann Anbor, 1975.

[7] C. Blum and A. Roli, "Metaheuristics in combinatorial optimization: Overview and conceptual comparison," ACM Comput. Surv., 35, 268-308, 2003.

[8] E.G. Talbi, Metaheuristics: From Design to Implementation, Wiley, 2009.

[9] X.S. Yang, Nature-Inspired Metaheuristic Algorithms, Luniver Press, 2008.

[10] X.S. Yang and S.Deb, "Engineering optimization by cuckoo search," Int. J. Math. Modelling & Num. Optimization, 1, 330-343, 2010.

[11] X.S. Yang, "A new metaheuristic bat-inspired algorithm," in: Nature Inspired Cooperative Strategies for Optimization. (Eds. J. R. Gonzalez et al.), Springer, SCI 284, 65-74 (2010b).

[12] A. Ayesh, "Swarm-based emotion modelling," Int. J. Bio-Inspired Computation, 118-124, 2009.

[13] R. Gross and M. Dorigo, "Towards group transport by swarms of robots," Int. J. Bio-Inspired Computation, 1, 1-13, 2009.

[14] J. Kennedy, R.C. Eberhart, Swarm intelligence, Academic Press, 2001.

[15] S. Kumar, V.K. Sharma, and R. Kumari. "A Novel Hybrid Crossover based Artificial Bee Colony Algorithm for Optimization Problem." International Journal of Computer Applications 82, 2013.

[16] S. Pandey and S. Kumar "Enhanced Artificial Bee Colony Algorithm and It's Application to Travelling Salesman Problem." HCTL Open International Journal of Technology Innovations and Research, Volume 2, 2013:137-146.

[17] J.C. Bansal, H. Sharma and S.S. Jadon "Artificial bee colony algorithm: a survey." International Journal of Advanced Intelligence Paradigms 5.1 (2013): 123-159.

[18] S. Das, A. Abraham, U.K. Chakraborty, and A. Konar, "Differential evolution using a neighborhood-based mutation operator." Evolutionary Computation, IEEE Transactions on, 13(3):526–553, 2009.

[19] R.C. Eberhart, Y. Shi, and J. Kennedy. Swarm intelligence. Elsevier, 2001.

[20] D.B. Fogel. "Introduction to evolutionary computation," Evolutionary Computation: Basic algorithms and operators, 1:1, 2000.

[21] L.J. Fogel, A.J. Owens, and M.J. Walsh. Artificial intelligence through simulated evolution. 1966.

[22] J.R. Koza. Genetic programming: A paradigm for genetically breeding populations of computer programs to solve problems. Citeseer, 1990.

[23] M.G.H. Omran, A.P. Engelbrecht, and A. Salman. "Differential evolution methods for unsupervised image classification." In Evolutionary Computation, 2005. The 2005 IEEE Congress on, volume 2, pages 966–973. IEEE, 2005.

[24] A. Ozturk, S. Cobanli, P. Erdogmus, and S. Tosun. "Reactive power optimization with artificial bee colony algorithm." Scientific Research and Essays, 5(19):2848–2857, 2010.

[25] K.V. Price, R.M. Storn, and J.A. Lampinen. "Differential evolution: a practical approach to global optimization," Springer Verlag, 2005.

[26] AK Qin, VL Huang, and PN Suganthan. Differential evolution algorithm with strategy adaptation for global numerical optimization. Evolutionary Computation, IEEE Transactions on, 13(2):398–417, 2009.

[27] I. Rechenberg. Cybernetic solution path of an experimental problem. Royal aircraft establishment, library translation 1122, 1965. reprinted in evolutionary computationthe fossil record, db fogel, ed., chap. 8, pp297-309, 1998.

[28] T. Rogalsky, S. Kocabiyik, and RW Derksen. Differential evolution in aerodynamic optimization. Canadian Aeronautics and Space Journal, 46(4):183–190, 2000.

[29] R. Storn and K. Price. Differential evolution-a simple and efficient adaptive scheme for global optimization over continuous spaces. International computer science institute - publications - TR, 1995.

[30] A. Baykasoglu, L. Ozbakir, and P. Tapkan, "Artificial bee colony algorithm and its application to generalized assignment problem," In Swarm Intelligence, Focus on Ant and Particle Swarm Optimization, ITech Education and Publishing, Vienna, Austria, pp. 113–144, 2007.

[31] D.Karaboga, "An idea based on bee swarm for numerical optimization," Tech. Rep. TR-06, Erciyes University，Engineering Faculty，Computer Engineering Department, 2005.

[32] D. Karaboga and B. Basturk. "A powerful and efficient algorithm for numerical function optimization: artificial bee colony (ABC) algorithm," J. Global Optim., vol. 39, pp. 459–471, 2007.

[33] D. Karaboga and B. Basturk. "On the performance of artificial bee colony (ABC) algorithm," Appl. Soft Comput., vol. 8, pp. 687–697, 2008.

[34] N. Karaboga, "A new design method based on artificial bee colony algorithm for digital IIR filters," J. Franklin I., vol. 346, pp. 328–348, May 2009.

[35] D. Karaboga, B. Akay, and C. Ozturk. "Artificial bee colony (ABC) optimization algorithm for training feed-forward neural networks." Modeling decisions for artificial intelligence. Springer Berlin Heidelberg, 2007. 318-329.

[36] A. Singh, "An artificial bee colony algorithm for the leafconstrained minimum spanning tree problem," Appl. Soft Comput., vol. 9, pp. 625–631, March 2009.

[37] R. S. Rao, S. V. L. Narasimham, and M. Ramalingaraju. "Optimization of distribution network configuration for loss reduction using artificial bee colony algorithm." International Journal of Electrical Power and Energy Systems Engineering 1.2 (2008): 116-122.

[38] M. Sonmez, "Artificial bee colony algorithm for optimization of truss structures." Applied Soft Computing 11.2 (2011): 2406-2418.

[39] F. Kang, J. Li, and Q. Xu, "Structural inverse analysis by hybrid simplex artificial bee colony algorithms," Comput. Struct., vol. 87, pp. 861-870, July 2009.

[40] D. Karaboga and B. Akay, "A comparative study of artificial bee colony algorithm". Applied Mathematics and Computation, 214(1):108–132, 2009.

[41] P. Moscato, "On evolution, search, optimization, genetic algorithms and martial arts: Towards memetic algorithms." Tech. Rep. Caltech Concurrent Computation Program, Report. 826, California Inst. of Tech., Pasadena, California, USA (1989).

[42] E. Burke, J. Newall, R. Weare.: A memetic algorithm for university exam timetabling. In: E. Burke, P. Ross (eds.) The Practice and Theory of Automated Timetabling, Lecture Notes in Computer Science, vol. 1153, pp. 241–250. Springer Verlag (1996).

[43] R. Carr, W. Hart, N. Krasnogor, E. Burke, J. Hirst, J. Smith, "Alignment of protein structures with a memetic evolutionary algorithm," In: Proceedings of the Genetic and Evolutionary Computation Conference. Morgan Kaufman (2002).

[44] R. Cheng, M. Gen, "Parallel machine scheduling problems using memetic algorithms." Computers & Industrial Engineering 33(3–4), 761–764 (1997).

[45] C. Fleurent, J. Ferland, "Genetic and hybrid algorithms for graph coloring." Annals of Operations Research 63, 437–461 (1997).

[46] G. Gutin, D. Karapetyan, N. Krasnogor, "Memetic algorithm for the generalized asymmetric traveling salesman problem." In: M. Pavone, G. Nicosia, D. Pelta, N. Krasnogor (eds.) Proceedings of the 2007Workshop On Nature Inspired Cooperative Strategies for Optimisation. Lecture Notes in Computer Science (LNCS), vol. to appear. Springer (2007)

[47] L. He, N. Mort, "Hybrid genetic algorithms for telecomunications network back-up routeing." BT Technology Journal 18(4) (2000)

[48] C. Reeves, "Hybrid genetic algorithms for bin-packing and related problems." Annals of Operations Research 63, 371–396 (1996)

[49] M. Tang, X. Yao, "A memetic algorithm for VLSI floorplanning." Systems, Man, and Cybernetics, Part B, IEEE Transactions on 37(1), 62–69 (2007). DOI 10.1109/TSMCB.2006.883268.

[50] W. E. Hart, "Adaptive Global Optimization with Local Search." Ph.D. Thesis, University of California, San Diego (1994)

[51] G. M. Morris, D. S. Goodsell, R. S. Halliday, R. Huey, W. E. Hart, R. K. Belew, A. J. Olson,: "Automated docking using a lamarkian genetic algorithm and an empirical binding free energy function." J Comp Chem 14, 1639–1662 (1998)

[52] H. Wang, D. Wang, and S. Yang. "A memetic algorithm with adaptive hill climbing strategy for dynamic optimization problems." Soft Computing 13.8-9 (2009): 763-780.

[53] D. Liu, K. C. Tan, C. K. Goh, W. K. Ho, "A multi-objective memetic algorithm based on particle swarm optimization." Systems, Man, and Cybernetics, Part B, IEEE Transactions on 37(1), 42–50 (2007).

[54] Y Wang, JK Hao, F Glover, Z Lü, "A tabu search based memetic algorithm for the maximum diversity problem." Engineering Applications of Artificial Intelligence 27 (2014): 103-114.

[55] X. Xue, Y. Wang, and A. Ren. "Optimizing ontology alignment through Memetic Algorithm based on Partial Reference Alignment." Expert Systems with Applications 41.7 (2014): 3213-3222.

[56] O. Chertov and Dan Tavrov. "Memetic Algorithm for Solving the Task of Providing Group Anonymity." Advance Trends in Soft Computing. Springer International Publishing, 2014. 281-292.

[57] J.C. Bansal, H. Sharma, K.V. Arya and A. Nagar, "Memetic search in artificial bee colony algorithm." Soft Computing (2013): 1-18.

[58] F Kang, J Li, Z Ma, H Li, "Artificial bee colony algorithm with local search for numerical optimization." Journal of Software 6.3 (2011): 490-497.

[59] I. Fister, I. Fister Jr, J. Bres, V. Zumer. "Memetic artificial bee colony algorithm for large-scale global optimization." Evolutionary Computation (CEC), 2012 IEEE Congress on. IEEE, 2012.

[60] R. Hooke and T.A. Jeeves. "Direct search solution of numerical and statistical problems." Journal of the ACM, 8(2):212–229, 1961.

[61] J. Kiefer, "Sequential minimax search for a maximum." In Proc. Amer. Math. Soc, volume 4, pages 502–506, 1953.

[62] M.M. Ali, C. Khompatraporn, and Z.B. Zabinsky. "A numerical evaluation of several stochastic algorithms on selected continuous global optimization test problems." Journal of Global Optimization, 31(4):635–672, 2005.

[63] P.N. Suganthan, N. Hansen, J.J. Liang, K. Deb, YP Chen, A. Auger, and S. Tiwari. "Problem definitions and evaluation criteria for the CEC 2005 special session on real-parameter optimization." In CEC 2005, 2005.

[64] G. Zhu and S. Kwong, "Gbest-guided artificial bee colony algorithm for numerical function optimization." Applied Mathematics and Computation, 217(7):3166–3173, 2010.

[65] A. Banharnsakun, T. Achalakul, and B. Sirinaovakul, "The best-so-far selection in artificial bee colony algorithm." Applied Soft Computing, 11(2):2888–2901, 2011.

[66] B. Akay and D. Karaboga. "A modified artificial bee colony algorithm for real-parameter optimization." Information Sciences, doi:10.1016/j.ins.2010.07.015, 2010.

[67] G.C. Onwubolu and B.V. Babu. "New optimization techniques in engineering." Springer Verlag, 2004.

[68] E. Sandgren, "Nonlinear integer and discrete programming in mechanical design optimization." Journal of Mechanical Design, 112:223, 1990.